# Hybrid Grey Interval Relation Decision-Making in Artistic Talent Evaluation of Player*

Gol Kim [a], Yunchol Jong [a], Sifeng Liu [b], Choe Rim Shong [c]

[a] Center of Natural Sciences, University of Sciences, Kwahakdong-1, Unjong District, Pyongyang, DPR Korea , E-mail: golkim124@yahoo.com

[b] College of Economics and Management, Nanjing University of Aeronautics and Astronautics, Nanjing, 210016, China, E-mail: sfliu@nuaa.edu.cn

[c] Pyongyang musical University, Pyongyang, DPR Korea

**Abstract** —-This paper proposes a grey interval relation TOPSIS method for the decision making in which all of the attribute weights and attribute values are given by the interval grey numbers. In this paper, all of the subjective and objective weights are obtained by interval grey number and decision-making is based on four methods such as the relative approach degree of grey TOPSIS, the relative approach degree of grey incidence and the relative approach degree method using the maximum entropy estimation using 2-dimensional Euclidean distance.

   A multiple attribute decision-making example for evaluation of artistic talent of Kayagum (stringed Korean harp) players is given to show practicability of the proposed approach.

## 1. Introduction

The multiple attribute decision-making (MADM) problems are of the most interesting problems for many decision-making experts. This problem arises in various fields of the real life, and constitutes very important content in scientific research such as management science, decision-making theory, system theory, operational research and economics.

Now, many effective methods to determine the attributive weights have been studied for MADM. Those are the subjective weight determining methods such as the feature vector method ( Saaty T.L. 1977 ), the least square sum method (Chu A Tw, Kalaba R E, Spingarn K, 1979), Delphi and AHP method (Hwang C.L., Lin M, 1987), and the objective weight determining methods such as the entropy method (Hwang C.L., Yoon K, 1981), the principal component analysis (Yan Jian-huo, 1989 ) and DEA (Data Envelopment Analysis) (Ye Chen, Kevin W. Li, Haiyan Xu and Sifeng Liu, 2009).

The final ranking method affects greatly on the decision-making process. Hwang and Yoon (1981) proposed a new approach, TOPSIS (Technique for Order Preference by Similarity to Ideal Solution) for solving MADM problem. Recently, TOPSIS methods with interval weights (Gao Feng-ji, et al, 2005) and multiple attribute interval number TOPSIS (Chu A Tw, Kalaba R E, Spingarn K, 1979) have been studied. Guo Kai-hong and Mu You-jing (2012) studied the relation between several possibility degree formulas and proposed a possibility degree matrices-based method that aimed to objectively determine the weights of criteria in MADM with intervals. A hybrid approach integrating OWA (Ordered Weighted Averaging) aggregation into TOPSIS is proposed to tackle multiple criteria decision analysis (MCDA) problems (Ye Chen, Kevin W. Li, Si-feng Liu, 2011). A hybrid approach of DEA (Data Envelopment Analysis) and TOPSIS is proposed for MCDA in emergency management (Ye Chen, Kevin W. Li, Haiyan Xu and Sifeng Liu, 2009).

---

* This work was supported in part by Nanjing University of Aeronautics and Astronautics, China.

The grey incidence degree in grey system theory is a very important technical conception. The computational formulas of incidence degree such as grey incidence degree, grey absolute incidence degree and grey comprehensive incidence degree are introduced and the concepts of grey relation decision-making are given (Liu Si-feng, Lin Yi, 2004). Luo Dang, Liu Si-feng et al (2005) extended the traditional grey relation decision-making method to interval grey number, proposed a choosing method of plan based on maximal degree and constructed a formula of grey interval incidence degree and a grey interval relative incidence degree. The ideal optimal plan for MADM problem was defined and a formula of grey interval incidence coefficient was obtained (Dang Yao-Guo, Liu Si-feng et al, 2004). Other methods in grey decision-making are the grey clustering decision-making (Mi Chuan-min, Liu Si-feng et al, 2006) and the grey incidence projection method (Zhang-Chao, et al, 2007).

This paper considers a hybrid MADM problem with interval attribute and interval decision matrix, and presents a grey interval relation method which considers comprehensive weight and preference of decision-making. First, the subjective weights of attributes are obtained as interval number based on group AHP method. Next, the first objective weights are determined based on optimization method and the second objective weights are obtained by interval number based on entropy method. Then, the comprehensive weights of attributes for decision-making are determined by combining the subjective weight and the objective weight using multiplicative composition method. Finally, three grey relation decision-making methods are proposed such as the evaluation of plan by the relative approach degree of grey TOPSIS, the evaluation by the relative approach degree of grey incidence, and the evaluation by the relative degree of grey incidence using maximum entropy method. The final rank based on rank vectors of three methods is obtained by the weighted Borda method. An example of MADM for artistic talent evaluation of Kayagum (a kind of Korean national musical instrument) players is given to show the advantage of our method.

## 2. Some concepts and normalization of decision matrix

**[Definition 1]** Let $a(\otimes) \in [\underline{a}, \overline{a}]$ and $b(\otimes) \in [\underline{b}, \overline{b}]$ be two interval grey numbers. Then, distance between $a(\otimes) \in [\underline{a}, \overline{a}]$ and $b(\otimes) \in [\underline{b}, \overline{b}]$ is defined by

$$d(a(\otimes), b(\otimes)) = \sqrt{(\overline{b} - \overline{a})^2 + (\underline{b} - \underline{a})^2} .$$

Let $A = \{A_1, A_2, \ldots, A_n\}$ be a set of the decision plans and $G = \{G_1, G_2, \ldots, G_m\}$ a set of attributes. The value of the attribute $G_j$ for plan $A_i$ is given the non-negative interval number by $a_{ij}(\otimes) \in [\underline{a}_{ij}, \overline{a}_{ij}]$, $(0 \leq \underline{a}_{ij}, \leq \overline{a}_{ij}, i = \overline{1,n}; j = \overline{1,m})$.

Let $a_i(\otimes) = (a_{i1}(\otimes), a_{i2}(\otimes), \cdots, a_{im}(\otimes))$, $i = 1, \cdots, n$ be attribute vector and $R(\otimes) = (a_{ij}(\otimes))_{n \times m}$ be decision matrix. The normalization of $a_{ij}(\otimes)$ is given as follows.

For the attribute of cost type,

$$\underline{x}_{ij}(\otimes) = \frac{1/\overline{a}_{ij}(\otimes)}{\sum_{i=1}^{n}(1/\underline{a}_{ij}(\otimes))}, \quad \overline{x}_{ij}(\otimes) = \frac{1/\underline{a}_{ij}(\otimes)}{\sum_{i=1}^{n}(1/\overline{a}_{ij}(\otimes))} \quad (i = \overline{1,n}, j = \overline{1,m}) .$$

For the attribute of effect type,

$$\underline{x}_{ij}(\otimes) = \frac{\underline{a}_{ij}(\otimes)}{\sum_{i=1}^{n}\overline{a}_{ij}(\otimes)}, \quad \overline{x}_{ij}(\otimes) = \frac{\overline{a}_{ij}(\otimes)}{\sum_{i=1}^{n}\underline{a}_{ij}(\otimes)} \quad (i = 1, \cdots, n, j = 1, \cdots, m)$$

**[Definition 2]** Let $X = (x_{ij}(\otimes))_{n \times m}$ be a normalized decision matrix. The attribute vector of each plan is $x_i(\otimes) = (x_{i1}(\otimes), x_{i2}(\otimes), \cdots, x_{im}(\otimes))$, $i = 1, \cdots, n$, where $x_{ij}(\otimes) \in [\underline{x}_{ij}, \overline{x}_{ij}]$ is non-negative interval grey number on $[0,1]$.

## 3. Determining of attribute weights

### 3.1. Subjective weight determining of attributes

Let $\alpha_l = [\alpha_l^1, \cdots, \alpha_l^j, \cdots, \alpha_l^m]$, ($l = \overline{1, L}$) be the attribute weights determined by AHP from the decision-making group. The weight of attribute $G_j$ is given as interval grey number $\alpha_j(\otimes) \in [\underline{\alpha}_j, \overline{\alpha}_j]$, $0 \leq \underline{\alpha}_j \leq \overline{\alpha}_j$, where $\underline{\alpha}_j = \min\limits_{1 \leq l \leq L}\{\alpha_l^j\}$, $\overline{\alpha}_j = \max\limits_{1 \leq l \leq L}\{\alpha_l^j\}$, $j = \overline{1, m}$.

### 3.2. Objective weight of attributes
#### 3.2.1. Objective weight determining by optimization

We define the deviation of decision plan $A_i$ from all other decision plans for attribute $G_j$ in normalized decision matrix $X = (x_{ij}(\otimes))_{n \times m}$ as follows

$$D_{ij}(\beta^{opt}) = \sum_{k=1}^{n} d(x_{ij}, x_{kj})\beta_j^{opt} = \sum_{k=1}^{n} \sqrt{(\overline{x}_{kj} - \overline{x}_{ij})^2 + (\underline{x}_{kj} - \underline{x}_{ij})^2}\, \beta_j^{opt}$$

In order to choose a proper weight vector $\beta^{opt}$ such that sum of overall deviation for the decision plan attains maximum, we define a deviation function such as

$$D(\beta) = \sum_{j=1}^{m}\sum_{i=1}^{n}\sum_{k=1}^{n} d(x_{ij}(\otimes), x_{kj}(\otimes))\beta_j$$

and solve the following nonlinear programming problem.

[P1]  $\max D(\beta) = \sum\limits_{j=1}^{m}\sum\limits_{i=1}^{n}\sum\limits_{k=1}^{n} d(x_{ij}(\otimes), x_{kj}(\otimes))\beta_j$,

$$s.t. \sum_{j=1}^{m} \beta_j^2 = 1,\ \beta_j \geq 0,\ j = 1, \cdots, m$$

**[Theorem 1]** The solution of problem P1 is given by

$$\overline{\beta}_j = \frac{\sum\limits_{i=1}^{n}\sum\limits_{k=1}^{n} d(x_{ij}(\otimes), x_{kj}(\otimes))}{\sqrt{\sum\limits_{j=1}^{m}\left[\sum\limits_{i=1}^{n}\sum\limits_{k=1}^{n} d(x_{ij}(\otimes), x_{kj}(\otimes))\right]^2}},\ j = 1, \cdots, m.\quad \square$$

By the normalization of $\overline{\beta}_j$, $j = 1, \cdots, m$, we obtain

$$\beta_j^{opt} = \frac{\sum\limits_{i=1}^{n}\sum\limits_{k=1}^{n} d(x_{ij}(\otimes), x_{kj}(\otimes))}{\sum\limits_{j=1}^{m}\sum\limits_{i=1}^{n}\sum\limits_{k=1}^{n} d(x_{ij}(\otimes), x_{kj}(\otimes))},\ j = 1, \cdots, m.$$

#### 3.2.2. Objective weight determining by entropy method
The entropy weights of the normalized decision matrix $X = (x_{ij}(\otimes))_{n \times m}$, $x_{ij}(\otimes) \in [\underline{x}_{ij}, \overline{x}_{ij}]$ for lower bound $\underline{x}_{ij}$ and upper bound $\overline{x}_{ij}$ of grey number $x_{ij}(\otimes)$ are determined as follows.
For lower bound $\underline{x}_{ij}$, letting

$$\underline{p}_{ij} = \frac{\underline{x}_{ij}}{\sum_{i=1}^{n} \underline{x}_{ij}}, \quad (i = \overline{1,n}, j = \overline{1,m}),$$

the entropy value of $j$ th attribute is given by $\underline{E}_j = -k \sum_{i=1}^{n} \underline{p}_{ij} \ln \underline{p}_{ij}$, $j = \overline{1,m}$, where $k = \dfrac{1}{\ln n}$. In the above formula, if $p_{ij} = 0$, then we regard that $p_{ij} \ln p_{ij} = 0$. Then $0 \le E_j \le 1$, $j = 1, \cdots, m$ and the deviation coefficient for $j$ th attribute is given by

$$\underline{\eta}_j = 1 - \underline{E}_j, \quad j = 1, \cdots, m.$$

The entropy weight $\underline{\beta}^{ent} = \left( \underline{\beta}_1^{ent}, \cdots, \underline{\beta}_j^{ent}, \cdots, \underline{\beta}_m^{ent} \right)$ for lower bound $\underline{x}_{ij}$ is given by

$$\underline{\beta}_j^{ent} = \frac{\underline{\eta}_j}{\sum_{j=1}^{m} \underline{\eta}_j} = \frac{1 - \underline{E}_j}{\sum_{j=1}^{m} (1 - \underline{E}_j)} = \frac{1 - \underline{E}_j}{m - \sum_{j=1}^{m} \underline{E}_j}, \quad j = \overline{1,m}.$$

Similarly, the entropy weight $\overline{\beta}^{ent} = \left( \overline{\beta}_1^{ent}, \cdots, \overline{\beta}_j^{ent}, \cdots, \overline{\beta}_m^{ent} \right)$ for upper bound $\overline{x}_{ij}$ is

$$\overline{\beta}_j^{ent} = \frac{1 - \overline{E}_j}{m - \sum_{j=1}^{m} \overline{E}_j}, \quad j = \overline{1,m},$$

where $\overline{E}_j = -k \sum_{i=1}^{n} \overline{p}_{ij} \ln \overline{p}_{ij}$ and $\overline{p}_{ij} = \dfrac{\overline{x}_{ij}}{\sum_{i=1}^{n} \overline{x}_{ij}}$ $(i = \overline{1,n}, j = \overline{1,m})$.

### 3.2.3. Determining of comprehensive objective weights

The comprehensive objective weight is determined by the interval grey number

$$\beta(\otimes) = (\beta_1(\otimes), \beta_2(\otimes), \cdots, \beta_m(\otimes)), \beta_j(\otimes) \in [\underline{\beta}_j, \overline{\beta}_j],$$

$$\underline{\beta}_j(\otimes) = \min \{\underline{\beta}_j^{opt}, \underline{\beta}_j^{ent}, \overline{\beta}_j^{ent}\}, \overline{\beta}_j(\otimes) = \max \{\beta_j^{opt}, \underline{\beta}_j^{ent}, \overline{\beta}_j^{ent}\}.$$

### 3.3. Determining of final comprehensive weights

The final comprehensive weight is determined by

$$w_j(\otimes) = \frac{\alpha_j(\otimes) \times \beta_j(\otimes)}{\sum_{j=1}^{m} \alpha_j(\otimes) \times \beta_j(\otimes)}, \quad j = \overline{1,m}$$

where $\alpha_j(\otimes)$ and $\beta_j(\otimes)$ are the subjective weight and the objective weight for $j$ th attribute, respectively. Thus, the weight of the attribute $G_j$ is a interval grey number $w_j(\otimes) \in [\underline{w}_j, \overline{w}_j]$ such as $0 \le \underline{w}_j \le \overline{w}_j \le 1$, $j = \overline{1,m}$.

## 4. Some evaluation methods of the decision plans

### 4.1. Evaluation of plan by the relative approach degree of grey TOPSIS method

Assume that the subjective preference value of the plan $A_i$ is given by the interval grey number $q_i(\otimes)$, where $q_i(\otimes) \in [\underline{q}_i, \overline{q}_i]$, $0 \le \underline{q}_i \le \overline{q}_i \le 1$, $i = \overline{1,n}$. The normalized decision matrix with the subjective preference is $\tilde{Z} = (z_{ij}(\otimes))_{n \times m}$, where $z_{ij}(\otimes) = \frac{1}{2} q_i(\otimes) + \frac{1}{2} x_{ij}(\otimes) \in \left[ \frac{1}{2} \underline{q}_i + \frac{1}{2} \underline{x}_{ij}, \frac{1}{2} \overline{q}_i + \frac{1}{2} \overline{x}_{ij} \right]$.

Let $\tilde{Y} = (y_{ij}(\otimes))_{n \times m}$ be the comprehensive weighted decision matrix such as
$$y_{ij}(\otimes) = w_j(\otimes) z_{ij}(\otimes) \in [\underline{y}_{ij}, \overline{y}_{ij}], \ i = \overline{1, n}, \ j = \overline{1, m}.$$

The attribute vector of each plan for the normalized comprehensive weighted decision matrix is $y_i(\otimes) = (y_{i1}(\otimes), y_{i2}(\otimes), \cdots, y_{im}(\otimes)), \ i = \overline{1, n}$.

**[Definition 3]** Let
$$\underline{y}_j^+ = \max_{1 \le i \le n} \{\underline{y}_{ij}\}, \ \overline{y}_j^+ = \max_{1 \le i \le n} \{\overline{y}_{ij}\}, \ \underline{y}_j^- = \min_{1 \le i \le n} \{\underline{y}_{ij}\}, \ \overline{y}_j^- = \min_{1 \le i \le n} \{\overline{y}_{ij}\}, \ j = \overline{1, m}.$$

Then, the $m$-dimension interval grey number vector $y^+(\otimes)$ ($y^-(\otimes)$) such as
$$y^+(\otimes) = (y_1^+(\otimes), y_2^+(\otimes), \cdots, y_j^+(\otimes), \cdots, y_m^+(\otimes))$$
$$(y^-(\otimes) = (y_1^-(\otimes), y_2^-(\otimes), \cdots, y_j^-(\otimes), \cdots, y_m^-(\otimes)))$$
is called a positive (negative) ideal plan attribute vector, where $y_j^+(\otimes) \in [\underline{y}_j^+, \overline{y}_j^+]$, $y_j^-(\otimes) \in [\underline{y}_j^-, \overline{y}_j^-]$, $j = \overline{1, m}$.

Euclidian distance between each plan attribute vector $y_i(\otimes)$ and the positive (or negative) ideal plan attribute vector $y^+(\otimes)$ (or $y^-(\otimes)$) is
$$D_i^+ = \sqrt{\sum_{j=1}^{m} \left[ (\overline{y}_{ij} - \overline{y}_j^+)^2 + (\underline{y}_{ij} - \underline{y}_j^+)^2 \right]}$$

or

$$D_i^- = \sqrt{\sum_{j=1}^{m} \left[ (\overline{y}_{ij} - \overline{y}_j^-)^2 + (\underline{y}_{ij} - \underline{y}_j^-)^2 \right]}.$$

The relative approach degree between each evaluation plan and the optimal plan is
$$C_i = \frac{D_i^-}{D_i^+ + D_i^-}, \ i = \overline{1, n}.$$

The best plan is one corresponding to the largest $C_i$.

## 4.2. Evaluation of plan by the relative approach degree of grey incidence

**[Definition 4]** Let $\{y_{ij}(\otimes)\}_{n \times m}$ be the normalized comprehensive weighted decision matrix and $y_j^+(\otimes)$ and $y_j^-(\otimes)$ be the positive and negative ideal plan attribute vector, respectively. We define
$$r_{ij}^+ = \frac{\min_i \min_j d(y_{ij}(\otimes), y_j^+(\otimes)) + \rho \max_i \max_j d(y_{ij}(\otimes), y_j^+(\otimes))}{d(y_{ij}(\otimes), y_j^+(\otimes)) + \rho \max_i \max_j d(y_{ij}(\otimes), y_j^+(\otimes))},$$

$$r_{ij}^- = \frac{\min_i \min_j d(y_{ij}(\otimes), y_j^-(\otimes)) + \rho \max_i \max_j d(y_{ij}(\otimes), y_j^-(\otimes))}{d(y_{ij}(\otimes), y_j^-(\otimes)) + \rho \max_i \max_j d(y_{ij}(\otimes), y_j^-(\otimes))}.$$

Then, $r_{ij}^+$ ($r_{ij}^-$) is called the coefficient of positive (negative) ideal grey interval incidence of $y_{ij}(\otimes)$ with respect to the positive ideal attribute value $y_j^+(\otimes)$ ($y_j^-(\otimes)$), where $\rho \in (0,1)$ and, generally, $\rho = 0.5$ is taken.

**[Definition 5]** The matrix $P^+ = \{r_{ij}^+\}_{n \times m}$ ($P^- = \{r_{ij}^-\}_{n \times m}$) is called a grey incidence coefficient matrix of the given plan with respect to the positive (negative) ideal plan.

**[Definition 6]** Let
$$G(y^+(\otimes), y_i(\otimes)) = \frac{1}{m}\sum_{j=1}^{m} r_{ij}^+, \quad G(y^-(\otimes), y_i(\otimes)) = \frac{1}{m}\sum_{j=1}^{m} r_{ij}^-, \quad i = 1,\cdots,n.$$

Then $G(y^+(\otimes), y_i(\otimes))$ ($G(y^-(\otimes), y_i(\otimes))$) is called a grey interval incidence degree of the comprehensive attribute vector for the plan $A_i$ with respect to the positive (negative) ideal plan attribute vector.

**[Theorem 2]** The grey interval incidence degrees $G(y^+(\otimes), y_i(\otimes))$ and $G(y^-(\otimes), y_i(\otimes))$ satisfy the four axioms of grey incidence degree (Sifeng Liu and Lin Y., 2004), i.e. normality, pair-symmetry, wholeness and closeness.

The grey incidence relative approach degree is defined by introducing the preference coefficients as follows.

$$C_i = \begin{cases} \dfrac{G(y^+(\otimes), y_i(\otimes)) \cdot \theta_+}{G(y^+(\otimes), y_i(\otimes)) \cdot \theta_+ + G(y^-(\otimes), y_i(\otimes)) \cdot \theta_-} ; & 0 < \theta_+ < 1,\ \theta_- < 1 \\ \\ G(y^+(\otimes), y_i(\otimes)) ; & \theta_+ = 1,\ \theta_- = 0 \end{cases},$$

where $\theta_+$ and $\theta_-$ are the preference coefficients, respectively. Generally, we regard as $\theta_+ > \theta_-$ and choose it so as to satisfy $0 < \theta_+ \leq 1, 0 < \theta_- \leq 1$ and $\theta_+ + \theta_- = 1$. When $\theta_+ = \theta_- = \dfrac{1}{2}$, it becomes the canonical formula for the grey incidence relative approach degree.

The best plan corresponds to the largest value among of the relative approach degree $C_i$.

**4.3. Evaluation of plan by the grey relation relative approach degree using maximum entropy estimation**

**[Definition 7]** Let $G(y^+(\otimes), y_i(\otimes))$ and $G(y^-(\otimes), y_i(\otimes))$ be the grey interval incidence degree for the plan $A_i$ with respect to the positive ideal plan and the negative ideal plan, respectively. We denote the weights of these two grey interval incidence degrees by $\beta_1$ and $\beta_2$, respectively, where $(\beta_1 + \beta_2 = 1,\ \beta_1, \beta_2 \geq 0)$. Then,

$$C_i^{''} = \beta_1 G(y^+(\otimes), y_i(\otimes)) + \beta_2[1 - G(y^-(\otimes), y_i(\otimes))] \quad (i = \overline{1,n})$$

is called a grey comprehensive incidence degree of the attribute vector $y_i$ of $i$th plan.

To obtain $\beta_1$ and $\beta_2$ by entropy method, we solve the following optimization problem

[P3] $\quad \max\{\sum_{i=1}^{n}[\beta_1 G(y^+, y_i) + \beta_2(1 - G(y^-, y_i))] - \sum_{j=1}^{2}\beta_j \ln \beta_j\}$

$\quad\quad s.t.\ \beta_1 + \beta_2 = 1,\ \beta_1 \geq 0, \beta_2 \geq 0$.

By solving this problem, we obtain the following weights.

$$\beta_1 = e^{\sum_{i=1}^{n}(G(y^+,y_i)+G(y^-,y_i)-1)}(1 + e^{\sum_{i=1}^{n}(G(y^+,y_i)+G(y^-,y_i))})^{-1},$$

$$\beta_2 = (1 + e^{\sum_{i=1}^{n}(G(y^+,y_i)+G(y^-,y_i))})^{-1}.$$

The best plan is one with the largest value of $C_i^*$.

The final rank is determined by the weighted Borda method using the rank vectors obtained from the above three methods.

## 5. MADM for evaluation of musical talent of Kayagum player

Kim Gol and Choe Rim Shong (2007) considered the grey comprehensive evaluation for artistic portray level of Korean flute players. In this paper, we considered a MADM problem for evaluation of artistic talent of Kayagum players in Pyongyang Musical University, DPR Korea. The evaluation group consisted of five music experts. The artistic talent of each player was evaluated according to the following criteria (attributes): clearness of sound ($G_1$), acoustic intensity ($G_2$), softness of sound ($G_3$), vibrato ($G_4$) and descriptive level ($G_5$). The group determined the rank of five selected Kayagum players by the grey relational decision-making method proposed in this paper. All of the above five attributes are effect type. Therefore, these attribute have scored values from 1 (worst) to 10 (best). The decision matrix is given in Table 1 which was obtained from the experts' group.

**Table 1**. Decision matrix

|       | $G_1$ | $G_2$ | $G_3$ | $G_4$ | $G_5$ |
|-------|-------|-------|-------|-------|-------|
| $A_1$ | [6,8] | [8,9] | [7,8] | [5,6] | [8,9] |
| $A_2$ | [7,9] | [5,7] | [6,7] | [7,8] | [7,9] |
| $A_3$ | [5,7] | [6,8] | [7,9] | [6,7] | [8,9] |
| $A_4$ | [6,7] | [7,8] | [6,9] | [5,6] | [7,8] |
| $A_5$ | [7,8] | [6,7] | [6,8] | [5,6] | [9,10]|

The subjective preference of decision-making group to the players was given by the grey interval number $q(\otimes) = ([6, 8], [5, 7], [5, 7], [5, 7], [6, 8])$.

The relative approach degree of grey TOPSIS is $C = (0.9938, 0.0461, 0.0298, 0.0273, 0.9663)$. Thus, the rank of players is such as $A_1 \succ A_5 \succ A_2 \succ A_3 \succ A_4$.

Next, we evaluated the players by the relative approach degree of grey incidence. For $\theta_+ = \theta_-$, $C' = (0.6802, 0.3305, 0.3289, 0.3263, 0.6760)$. Thus, we obtain the rank such as $A_1 \succ A_5 \succ A_2 \succ A_3 \succ A_4$.

Then, the grey incidence relative approach degree using the maximum entropy is $C'' = (0.9435, 0.5210, 0.5215, 0.5199, 0.9294)$. Thus, the rank of plans is such as $A_1 \succ A_5 \succ A_3 \succ A_2 \succ A_4$.

The final rank by the weighted fuzzy Borda method is $A_1 \succ A_5 \succ A_2 \succ A_3 \succ A_4$.

## 6. Conclusion

In this paper, for MADM in which all of attribute weights and attribute values are given by interval grey number, we have proposed an interval weight determining method and three methods of grey interval relation decision-making: the evaluation of plans by the relative approach degree of grey TOPSIS method, the evaluation by the relative approach degree of grey incidence and the evaluation by the relative approach degree of grey incidence using maximum entropy estimation. The final rank of plans has been obtained by weighted Borda method considering the above three ranking results.

Our method consists of three stages. The first is finding of the subjective grey interval weights by group AHP, finding of the objective grey interval weights by optimization and entropy method, and then finding of the final grey interval weights by multiplicative composition using the grey interval subjective and objective weights. The second is to obtain the weighted grey interval decision matrix considering the comprehensive grey interval weights determined in the preceding steps for MADM with interval decision matrix. The third is that decision-making based on the relative approach degree of grey incidence, the relative approach degree of grey TOPSIS and the relative approach degree of grey incidence using maximum entropy estimation. The weighted Borda method is used for combining the results of three methods. The proposed method was applied to MADM problem for selecting the best Kayagum player in DPR Korea.